\documentclass[letterpaper]{article} 
\usepackage{aaai24}  
\usepackage{times}  
\usepackage{helvet}  
\usepackage{courier}  
\usepackage[hyphens]{url}  
\usepackage{graphicx} 
\usepackage{natbib}  
\usepackage{caption} 
\usepackage{algorithm}
\usepackage{algorithmic}

\usepackage{multirow}
\usepackage{amsthm}
\usepackage{booktabs}
\usepackage{makecell}
\usepackage{stackengine}
\usepackage{caption}
\usepackage{subcaption}
\usepackage{amsmath,amssymb,amsfonts}
\usepackage{bm}
\usepackage{color}
\usepackage{graphicx}
\usepackage{xcolor}
\usepackage{newfloat}
\usepackage{listings}
\DeclareCaptionStyle{ruled}{labelfont=normalfont,labelsep=colon,strut=off} 
\floatstyle{ruled}
\newfloat{listing}{tb}{lst}{}
\floatname{listing}{Listing}
\title{CLIP-guided Federated Learning on Heterogeneous and Long-Tailed Data}
\author{
    Jiangming Shi\textsuperscript{\rm 1}, Shanshan Zheng\textsuperscript{\rm 2},  Xiangbo Yin\textsuperscript{\rm 2}, Yang Lu\textsuperscript{\rm 2}, Yuan Xie\textsuperscript{\rm 3,4$^*$}, Yanyun Qu\textsuperscript{\rm 1,2\thanks{Corresponding author.}}
}
\affiliations{

  \textsuperscript{\rm 1} Institute of Artificial Intelligence,
  \textsuperscript{\rm 2} School of Informatics, Xiamen University\\
  \textsuperscript{\rm 3} East China Normal University, 
  \textsuperscript{\rm 4} Chongqing Institute of East China Normal University
  \\
{\tt\small  jiangming.shi@outlook.com, shanshanzheng@stu.xmu.edu.cn, S\_yinxb@163.com}\\
{\tt\small luyang@xmu.edu.cn, yxie@cs.ecnu.edu.cn, yyqu@xmu.edu.cn}\\
}

\usepackage{bibentry}

\begin{document}

\maketitle

\begin{abstract}
Federated learning (FL) provides a decentralized machine learning paradigm where a server collaborates with a group of clients to learn a global model without accessing the clients' data. User heterogeneity is a significant challenge for FL, which together with the class-distribution imbalance further enhances the difficulty of FL. Great progress has been made in large vision-language models, such as Contrastive Language-Image Pre-training (CLIP), which paves a new way for image classification and object recognition. Inspired by the success of CLIP on few-shot and zero-shot learning, we use CLIP to optimize the federated learning between server and client models under its vision-language supervision. It is promising to mitigate the user heterogeneity and class-distribution balance due to the powerful cross-modality representation and rich open-vocabulary prior knowledge. In this paper, we propose the CLIP-guided FL (CLIP2FL) method on heterogeneous and long-tailed data. In CLIP2FL, the knowledge of the off-the-shelf CLIP model is transferred to the client-server models, and a bridge is built between the client and server. Specifically, for client-side learning, knowledge distillation is conducted between client models and CLIP to improve the ability of client-side feature representation. For server-side learning, in order to mitigate the heterogeneity and class-distribution imbalance, we generate federated features to retrain the server model. A prototype contrastive learning with the supervision of the text encoder of CLIP is introduced to generate federated features depending on the client-side gradients, and they are used to retrain a balanced server classifier. Extensive experimental results on several benchmarks demonstrate that CLIP2FL achieves impressive performance and effectively deals with data heterogeneity and long-tail distribution. Code is available at \url{https://github.com/shijiangming1/CLIP2FL}.
\end{abstract}

\section{Introduction}

With the demand for data privacy protection, federated learning~\cite{PPDL,FLSurveyandBenchmarkforGenRobFair_arXiv23}~(FL) springs up, which only uses the local client models without their private data to train the server model, and it provides a distributed machine-learning paradigm with the communication efficiency and privacy protection. It allows users to collectively reap the benefits of a server model trained from client-side privacy data without accessing the client-side data~\cite{FedBN,FCCLPlus_TPAMI23}. 

Accompanied by the prospect of FL, it faces the challenge of data heterogeneity, which is a universal issue arising from disparities in data distribution among clients due to data source differences between clients~\cite{SCAFFOLD,FCCL_CVPR22}. Furthermore, real-world data frequently exhibits the long-tailed phenomenon~\cite{BFL}, that is, the class distribution is imbalanced, where the head classes have a large number of samples while the tail classes have a small number of samples. Long-tailed data makes client-side models biased toward the head class~\cite{shang2022federated}. It is known that the server model based on client models is destined to not work well if the client models are poor in performance. Thus, heterogeneity and long-tailed data make FL more challenging.

\begin{figure}[t]
  \centering
  \includegraphics[width=1.0\linewidth]{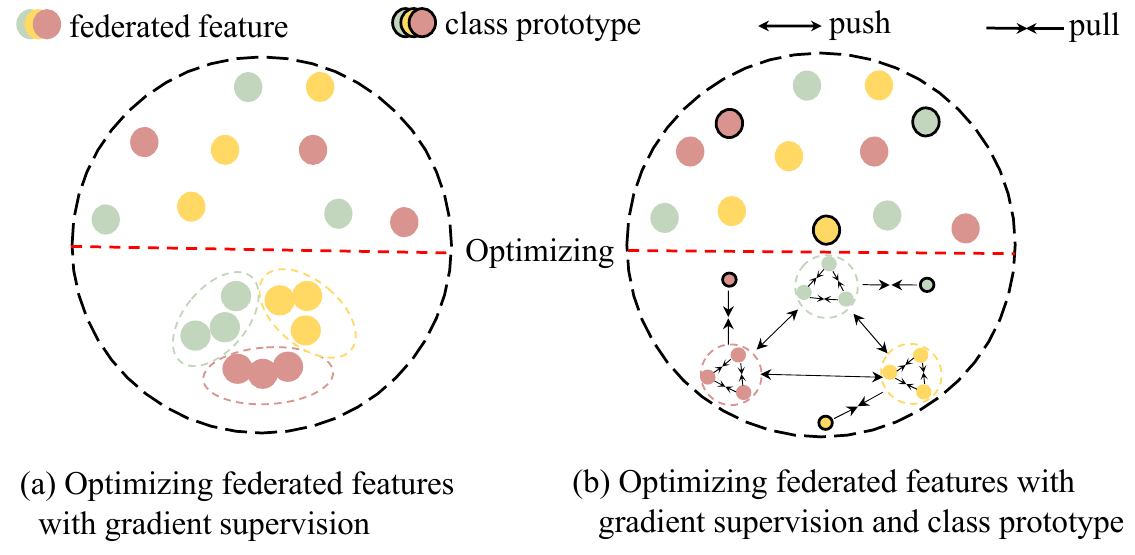}
  \caption{The figure illustrates the differences between CReFF and CLIP2FL: CReFF uses only gradients for supervision, while CLIP2FL introduces the textual features of the CLIP as prototypes for federated feature learning.}
  \label{fig:introduction}
\end{figure}

There are some FL methods for imbalanced data. In the works~\cite{duan2020self,yang2021federated}, the class distribution is appended to adopt client selection to match clients. Because clients need to upload the local class distribution to the server, the operation violates the principle of privacy protection of FL. In FedNova~\cite{wang2021addressing}, the auxiliary balanced data on the server are used to compute the class distribution output by the global class, but they are unavailable in the real world. CReFF~\cite{shang2022federated} is the latest solution, whose core idea is to implement the decoupling strategy to generate the class-distribution balanced federated features for the server model and to retrain the classifier by the federated features. However, in CReFF, the generation of federated features via client-side gradient information brings two limitations: 1) It is a one-to-many mapping between gradients and samples, which results in the problem becoming ill-posed; 2) It lacks semantic supervision, which could result in the federated features lacking discriminative ability for their respective classes. As shown in Figure \ref{fig:introduction}, we give the comparison of CReFF and CLIP2FL in designation. 

In short, the essential challenges of FL are the distinctiveness of feature representation and effective semantic supervision, but the existing methods have limitations in the solution of the two problems.
Recently, large visual-language models with open-vocabulary supervision have made great progress, such as Contrastive Vision-Language Pre-training (CLIP). It paves a new way for image classification and object recognition~\cite{CLIP2Scene}. CLIP is trained on large-scale free-available image-text pairs from websites and built the vision-language correlation to achieve promising open-vocabulary recognition. It is proven that CLIP benefits the few-shot and zero-shot learning~\cite{HWCLIP}. However, the potential of employing CLIP to enhance learning in FL remains unexplored until now. 

Considering that CLIP has rich prior knowledge and powerful vision-language cross-modality representation, we discuss how to use CLIP to solve the FL on heterogeneous and long-tailed data. There are two important issues: 1) How to use CLIP to improve the feature representation of client models; 2) How to use CLIP to mitigate the influence of heterogeneity and class-distribution imbalance on the server model. 
For the first issue, CLIP is used as ``Teacher" while the client model is treated as ``Student", and knowledge is transferred from Teacher to Student for improving the ability of feature representation. For the second issue, inspired by the success of the retraining strategy on class-distribution-imbalance visual classification, we generate the federated features and use them to retrain the balance server model. We use CLIP to constrain the generation of federated features from the client-side gradients under its effective semantic supervision, so the generated federated features may avoid heterogeneity and class-distribution imbalance to some degree. Moreover, the federate features are related to explicit semantic cues.  

Therefore, we propose the CLIP-guided federated learning method named CLIP2FL for heterogeneous and long-tailed data. CLIP2FL bridges client-side training with server-side training in FL via semantic supervision of CLIP. Firstly, in the client side, we conduct knowledge distillation to transfer the prior knowledge of an off-the-shelf CLIP model to local client models. The image and text encoders of CLIP are treated as the teacher model (Teacher) and the local client models are treated as the student model (Student). The class distribution output by ``Student" is enforced to be consistent with the counterpart by CLIP. CLIP2FL significantly mitigates the bias of client-side classifiers to the head classes. Secondly, contrastive learning is implemented to generate the federated features for the server model, which are used for retraining the server classifier. The text encoder of CLIP presents the prototype feature of a class, and the generated federated features are enforced to mimic the output features of the text encoder of CLIP. 

To summarize, our contributions are threefold:
\begin{itemize}
	\item We propose the CLIP-guided federated learning method (CLIP2FL) for heterogeneous and long-tailed data, which constructs a bridge between client and server training with CLIP supervision. To the best of our knowledge, we are the first to explore how CLIP benefits FL. 
        \item In the client side, we conduct knowledge distillation that transfers the vision-language prior knowledge of CLIP to local client models to improve the feature representation of client models. In the server side, unlike CReFF which generates the federated features only depending on the gradient information, we employ a prototype contrastive learning to make the federated features similar to the output features of the CLIP text encoder without accessing client data. Retraining the server model on federated features effectively mitigates the heterogeneity and class-distribution imbalance. 
	\item Extensive experimental results demonstrate that CLIP2FL is effective and efficient and it is superior to the state-of-the-art FL methods on multiple datasets.
\end{itemize}

\section{Related Work}
\subsection{Federated Learning with Heterogeneous Data}
The problem of data heterogeneity in FL is mainly caused by the fact that the data of the clients participating in the training are independently distributed and their sample distributions are not consistent, which leads to a serious decline in model accuracy. At present, there are many methods to solve this problem. They are divided into three categories: jointly improving the client and server models, stabilizing the local models, and improving the server model. The typical first category methods~\cite{PVR, DBLP:conf/ijcai/HuangS0L21,FPL_CVPR23,Ins} mainly apply optimization strategies to improve the optimization of the client and server. In the second category methods~\cite{FT-pFL, FedHealth, FedFTG, FMA,PR}, knowledge transfer is used to transfer the local knowledge in a model-agnostic to solve data heterogeneity. The third category methods~\cite{lin2020ensemble,chen2020fedbe} aim at eliminating the differences of heterogeneous data by improving model aggregation on the server. Although the existing methods mitigate the data heterogeneity to a certain extent, they do not consider the FL on long-tailed data.
\begin{figure*}[t]
	\centering
	\includegraphics[width=1.0\linewidth]{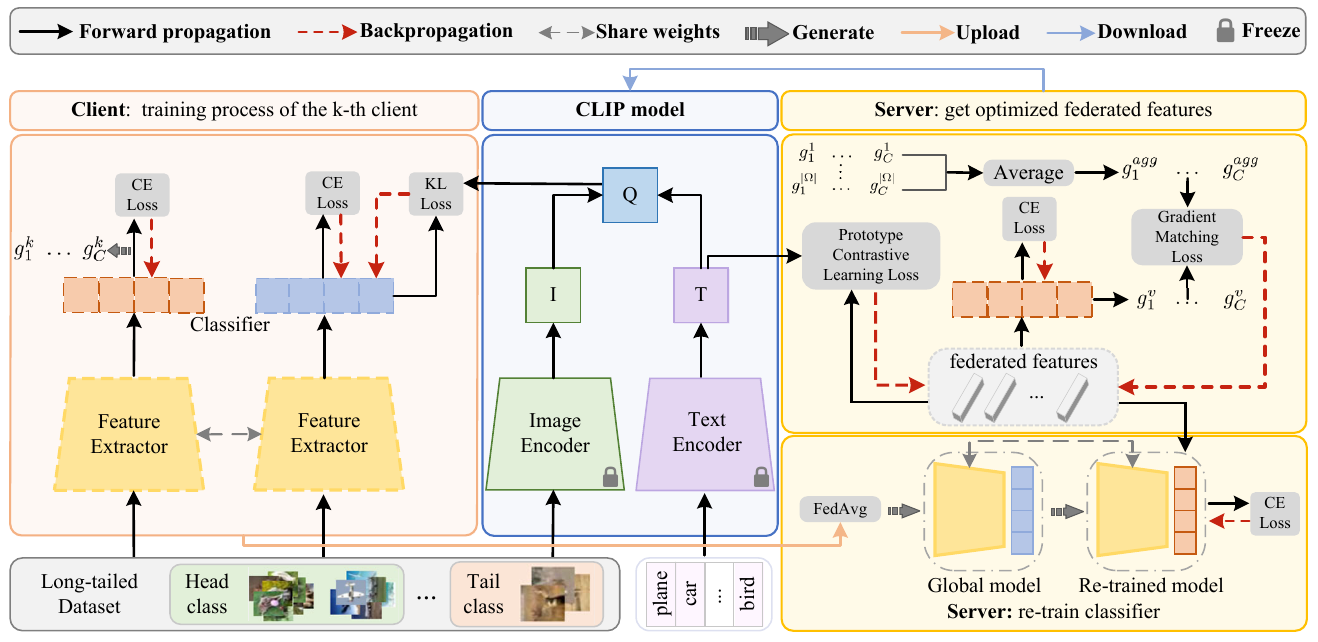} 
	\caption{The framework of CLIP2FL. It includes two core components: local training on clients and classifier re-training on the server. A prior knowledge-rich CLIP model acts as a bridge to connect the two components and helps the two core components to learn better.}
	\label{overframework}
\end{figure*}

\subsection{Contrastive Language-Image Pre-training}
In recent years, Vision-Language models have achieved advances, and Contrastive Language-Image Pre-training (CLIP) is a typical model. CLIP \cite{radford2021learning} is pre-trained on web-scale image-text data and shows unprecedented generality. CLIP is applied to improve many visual tasks, such as semantic segmentation (LSeg \cite{LSeg}, GroupViT \cite{GroupViT}), object detection (ViLD \cite{ViLD}, GLIP \cite{GLIP}), image generation (CLIPasso \cite{CLIPasso}), image retrieval (Context-I2W \cite{CIW}, AlignCMSS \cite{ABS}) and video understanding (ActionCLIP \cite{ActionCLIP}, CLIP4Clip \cite{CLIP4Clip}). It is recognized that CLIP benefits for unseen-class classification and data deficiency \cite{HRIS, ZeroCap}. However, it is unexplored how CLIP benefits federated learning. Taking inspiration from the accomplishments of CLIP in few-shot and zero-shot learning, we delve into how CLIP improves FL on heterogeneous and long-tailed data.

\section{Method}
\subsection{Problem Formulation and Notation}
 We aim to learn a global server model for image classification on heterogeneous and long-tailed data. CLIP2FL includes a central server model and $K$ client models. The local-client datasets are denoted by $\mathcal D^1$, $\mathcal D^2$, \ldots, $\mathcal D^K$, and the merge of local-client datasets is denoted by $\mathcal D = \cup_k  \mathcal D^k$, where $k \in \{1,2, \dots, K \}$. In this paper, the dataset $\mathcal D$ comprises $C$ classes and the class distribution \textbf{$\chi$} of $\mathcal D$  is long-tailed. Let $\mathcal D^k_c$ =$\left\{(x, y) \in \mathcal{D}^{k}: y=c\right\}$ denotes the set of samples with ground-truth label $c$ on the $k$-th client and contains $n^k_c$ samples, so the number of samples in the $c$-th class is $n_{c}=\sum_{k=1}^{K} n_{c}^{k}$. We rearrange them according to the number of their samples in descending order as $n_1$ $>$ $\ldots$  $>$ $n_c$  $>$ $\ldots$  $>$ $n_C$. 
In CLIP2FL, both the local client model and server model are classification models and contain two main components: a feature extractor $f_{\theta}$ w.r.t. the parameter $\theta$, and a linear classifier $g_{\varphi}$ w.r.t. the parameter $\varphi$. An input sample $x$ is fed into a classification model, and it is processed by a feature extractor that outputs a $d$-dimension feature represented as $z=f_{\theta}(x)$ and then it is processed by a classifier that outputs a logit vector $p$ = $g_{\varphi}(z)$ as the label prediction for  $x$. Here we define the classification network with the parameter $w=\{\theta, \varphi\}$. 

\subsection{Framework of CLIP2FL}
The common streamline of FL contains three steps in communication iteration: 1) The server broadcasts the global model to clients; 2) The client-side models are updated by training on their local data; 3) The client-side updated models are aggregated to the server, based on which a new global model is updated, and repeated until convergence.

Following the three steps of FL, CLIP2FL focuses on the latter two steps: the client-side learning and the server-side learning, as shown in Figure \ref{overframework}. For the client-side learning, we employ knowledge distillation to transfer the knowledge from CLIP to the local client models, enhancing the ability of feature representation of the client-side models. Thanks to the inherent vision-language correlation within CLIP, the problem of strong bias is alleviated after conducting knowledge distillation. As a result, the gradients of client models become more distinct and benefit the effectiveness of the ensuing server-side learning. For the server-side learning, CLIP2FL first generates federated features, which are used to retrain a balanced classifier of the server model. The class distributions of client-side private data are very different, e.g. some private data are long-tailed, and some private data only contain part of the classes, so it is impossible for the classifier of the server model to work well on heterogeneous and long-tailed data only with one-time iteration. It is necessary to retrain the classifier of the server model. The generated federated features make the server-side model less biased to the head classes based on the retraining strategy.

\subsection{Client-side Learning}\label{sectionlabel}
In this subsection, we introduce how CLIP benefits the local client models. Concretely, CLIP is treated as ``Teacher", while a local client model is treated as ``Student". CLIP offers guidance to the client models during each iteration, facilitating its updates and enhancements.
We directly use a ready-made pre-training CLIP while not training it again. CLIP typically contains an image encoder $I_{enc}$ whose input is an image and a text encoder $T_{enc}$ whose input is the text.  Given an input image ${x}$ with its label name ${l}$, we can get the visual feature $f_{i}^{'}$ = $I_{enc}$(${x}$) $\in {R}^{d_{i}}$  and the text feature $f_{r}^{'}$ = $T_{enc}$(${l}$) $\in {R}^{d_{r}}$. The class label is transformed into a sentence, e.g. ``cat" is transformed into ``This is a cat". 

In order to mitigate the heterogeneous and long-tailed problem, the output of client models is enforced to be consistent with the output of CLIP. The loss of local training is formulated as:
\begin{equation}
\label{local_training_loss}
\begin{aligned}
{L_{loc}} &= {L_{ce}}(y, p^t_k) + {\beta} \cdot KL(q^t_k || p^t_k),
\end{aligned}
\end{equation}
where ${L_{ce}(\cdot,\cdot)}$ is the cross-entropy loss, ${KL}$ is the  Kullback-Leibler divergence~\cite{joyce2011kullback} and $\beta$ is a hyper-parameter balanced the cross-entropy loss and the loss of knowledge distillation. The latter loss helps to narrow the output class prediction of the client and CLIP,  $p^t_k$ and  $q^t_k$ denote the output logits vector of the client $k$ and CLIP on local data $\mathcal D^k$ in the $t$th round, respectively. Then the $k$th client updates the received model similar to FedAvg~\cite{mcmahan2017communication} as follows:
\begin{equation}
\label{update_local_model}
\begin{aligned}
{w}_{k}^{t+1} \leftarrow {w}_{k}^{t}-\eta \nabla_{{w}} L_{loc}\left({w}^{t} ; \mathcal{D}^{k}\right).
\end{aligned}
\end{equation}

\begin{table*}[htb]
	\caption{Top-1 classification accuracy(\%) on CIFAR-10-LT and CIFAR-100-LT datasets with different FL methods, where the results are referred in \protect\cite{shang2022federated}. The best results are marked in bold.}
	\label{CIFAR10-100}
    \centering
	\begin{tabular}{llcccccc}
		\hline \multirow{2}{*}{{\textbf{Type }}} & \multirow{2}{*}{{\textbf{Method}}} & \multicolumn{3}{c}{\textbf{CIFAR-10-LT}} & \multicolumn{3}{c}{\textbf{CIFAR-100-LT}} \\
		\cline { 3 - 8 } & & IF=100 & IF=50 & IF=10 & IF=100 & IF=50 & IF=10 \\
		\hline \multirow{7}{*}{\makecell[l]{{\text { Heterogeneity-oriented }} \\ \text { FL methods }}} & 
		\text { FedAvg } & 56.17 & 59.36 & 77.45 & 30.34 & 36.35 & 45.87 \\
		& \text { FedAvgM } & 52.03 & 57.11 & 70.81 & 30.80 & 35.33 & 44.66 \\
		& \text { FedProx } & 56.92 & 60.89 & 76.53 & 31.67 & 36.30 & 46.10 \\
		& \text { FedDF } & 55.15 & 58.74 & 76.51 & 31.43 & 36.22 & 46.19 \\
		& \text { FedBE } & 55.79 & 59.55 & 77.78 & 31.97 & 36.39 & 46.25 \\
		& \text { CCVR } & 69.53 & 71.89 & 78.48 & 33.43 & 36.98 & 46.88 \\
		& \text { FedNova } & 57.79 & 63.91 & 77.79 & 32.64 & 36.62 & 46.75 \\
		\hline \multirow{3}{*}{\makecell[l]{{\text { Imbalance-oriented }} \\ \text { FL methods }}} & \text { Fed-Focal Loss } & 53.83 & 57.42 & 73.74 & 30.67 & 35.25 & 45.52 \\
		& \text { Ratio Loss } & 59.75 & 64.77 & 78.14 & 32.95 & 36.88 & 46.79 \\
		& \text { FedAvg+$\tau$-norm} & 49.95 & 51.41 & 72.08 & 26.22 & 33.71 & 43.65 \\
		\hline \text { SOTA } & \text { CReFF } & 70.55 & 73.08 & 80.71 & 34.67 & 37.64  & 47.08 \\
		\hline
		\hline \multirow{2}{*}{\text { Proposed method }} & \multirow{2}{*}{\text { CLIP2FL }}  & \textbf{73.37} & \textbf{75.35} & \textbf{81.18}  & \textbf{37.56} & \textbf{41.29}  & \textbf{48.20}  \\
		&\text {$\quad\quad$ } & ($ \textcolor{red}{\uparrow 2.82}$) & ($ \textcolor{red}{\uparrow 2.27}$) & ($ \textcolor{red}{\uparrow 0.47}$) & ($ \textcolor{red}{\uparrow 2.89}$)& ($ \textcolor{red}{\uparrow 3.65}$)  & ($ \textcolor{red}{\uparrow 1.12}$) \\
		\hline	
	\end{tabular}
\end{table*}

\subsection{Server-side Learning}
In this subsection, we introduce the server-side learning. Server-side learning comprises three important components: aggregation of the gradients of the client-side classifiers corresponding to their client-side models ${\hat{\varphi}}^t$, the generation of federated features, and retraining the global model. \\
\textbf{Gradient Aggregation.}
We compute the gradients of the current server classifier ${\hat{\varphi}}^t$, which is used to optimize the federated feature for re-training a balanced classifier on the server. Specifically, the current server classifier ${\hat{\varphi}}^t$ replaces the classifier of the $k$th client model, and with the resampled data from the dataset $\mathcal D_i$, the new client model produces $d$-dimension real sample features $\mathcal{Z}_{c}^{k}=\left.\left\{{z}_{c, i}^{k}\right\}\right|_{i=1} ^{n_{c}^{k}}$ for the $c$th class. Then we use the server classifier ${\hat{\varphi}}^t$ to compute the gradients $g_c^k \in {R}^{C \times d}$ of the $c$th class as below:
\begin{equation}
\begin{aligned}
\label{compute_real_gradient}    
{g}_{c}^{k}=\frac{1}{n_{c}^{k}} \sum_{i=1}^{n_{c}^{k}} \nabla_{\hat{\varphi}^t} L_{ce}\left({z}_{c, i}^{k}, y_{i}\right).
\end{aligned}
\end{equation}

After that, the real gradients $\left\{g_c^k \mid c \in {C}^{k}\right\}$ are uploaded to the server. Due to data heterogeneity, ${C}^{k}$ here indicates that only the class index set is involved in the current client training data. Importantly, CLIP2FL only obtains gradients subsequent to the irreversible averaging operation, without any direct access to the client data.\\
\textbf{Federated Feature Generation.}
Firstly, the server aggregates real gradients of each class and computes the average of the real gradients over all selected clients in $\Omega^t$:
\begin{equation}
\begin{aligned}
\label{aggregate_real_gradient}
{g}_{c}^{agg}=\frac{1}{\left|\Omega_{c}^{t}\right|} \sum_{k=1}^{\left|\Omega_{c}^{t}\right|} {g}_{c}^{k},
\end{aligned}
\end{equation}
where $\Omega_{c}^{t}$ is the subset of clients including  the $c$th class. 

After that, we generate a set of $d$-dimension federated features of size $m$ for each class in the $t$-th round, denoted as ${V}_{c}^{t}$ = $\left.\left\{{v}_{c, i}^{t}\right\}\right|_{i=1}^{m}$. The federated features are generated by enforcing the two constraints: 1) The gradient computed by the server classifier ${\hat{\varphi}}^t$ on the federated features should be consistent with the gradient computed by the real features; 2) The federated features should be similar to the class prototypes that are produced by the text encoder of CLIP. Due to highly distinctive class representations, we consider that prototypes are helpful to mitigate the strong bias problem.

As for the first constraint, we use the classifier ${\hat{\varphi}}^t$ to generate the gradient of the corresponding class for the federated features:

\begin{equation}
\label{compute_virtual_gradient}
\begin{aligned}
{g}_{c}^{{v}}=\frac{1}{m} \sum_{i=1}^{m} \nabla_{\hat{\varphi}^t} L_{ce}\left({v}_{c, i}^{t}, y_{i}\right).
\end{aligned}
\end{equation}

To further ensure consistency between federated features and real features, the gradient matching loss~\cite{zhao2021dataset} is used to measure the difference between the gradients generated by the server classifier for the two types of features and minimize it:
\begin{equation}
\label{loss_grad}
\begin{aligned}
{L_{grad}} = D\left({g}_{c}^{{v}}, {g}_{c}^{{agg}}\right)=\frac{1}{C} \sum_{j=1}^{C}\left(1-\frac{{g}_{c}^{{v}}[j] \cdot {g}_{c}^{{agg}}[j]}{\left\|{g}_{c}^{{v}}[j]\right\| \times \left\|{g}_{c}^{{agg}}[j]\right\|}\right),
\end{aligned}
\end{equation}
where ${g}[j]$ denotes the $j$th row of the gradient. This helps to make the federated features as similar as possible to the real features.

As for the second constraint, we use the prototype output by the text encoder of CLIP with strong semantic information to guide the generation of federated features. A prototype with strong supervision information benefits the generation of distinctive federated features. 
Concretely, we conduct prototype contrastive learning. The federated features ${V}_{c}^{t}$ and their intra-class prototype $f_{r}^c$ are treated as positive pair-wise samples while federated features ${V}_{c}^{t}$ and their inter-class federated features are treated as negative pair-wise samples. The loss of prototype contrastive learning is:
\begin{equation}
\begin{aligned}
\label{loss_ctra}
    {L_{pcl}} = \sum_{i=1}^{C \times m} {-\log \frac{\exp \left(\left \langle {{v}_{c, i}, f_{r}^c } \right \rangle / \tau\right)}{\sum_{j=1}^{C \times m} {{1}}_{[j \neq i]} \exp \left(\left \langle {v}_{c, i}, {v}_{j} \right \rangle / \tau\right)}},
\end{aligned}
\end{equation}

 where $\left \langle \cdot, \cdot \right \rangle$ denotes cosine similarity and $\tau$ is a temperature hyper-parameter. The total loss for obtaining optimized federated features is as follows:  
 \begin{equation}
\label{optimition_virtual_feature_total_loss}
\begin{aligned}
{L_{total}} &= {L_{grad}} + {\eta} \cdot {L_{pcl}},
\end{aligned}
\end{equation}
where $\eta$ is a hyper-parameter controlling the loss weights.

After the federated features ${V}^{t}$ have been optimized using the total loss function, we obtain an updated ${V}^{t+1}$ for retraining the classifier. 
Applying the two constraints to the generation of federated features provides two significant advantages. On the one hand, CReFF~\cite{shang2022federated} only uses gradient information, which is not enough to guide the generation of federated features (Eq. \ref{loss_grad}). Due to privacy protection, the gradient generated by the client is uploaded after an irreversible average operation, so the averaged gradient information and real features are one-to-many, which means the federated features are not very distinguishable, which is not conducive to subsequent classifier re-training. On the other hand, we utilize the CLIP model, which has rich prior knowledge, as an auxiliary module to assist in the training of federated features on the server. This allows the use of the rich semantic information contained within the CLIP model to guide the generation of federated features, resulting in more accurate and useful federated features. This neatly ties into the training of the client and shows how to make good use of vision-language models for FL. \\
\textbf{Classifier Re-Training.} After the local client model is updated,in the $t$th round,  clients in the selected set $\Omega^t$ upload their updated models ${w}_{k}^{t+1}$ to the server. The average of parameters of selected client models is assigned to the global model. The server model performs a weighted average to update the global model for the $t+1$ round where the weight of a class is its proportion to the amount of the whole dataset:
\begin{equation}
\begin{aligned}
\label{update_global_model}
{w}^{t+1}=\sum_{k \in \Omega^{t}} \frac{\left|\mathcal{D}^{k}\right|}{\sum_{k \in \Omega^{t}}\left|\mathcal{D}^{k}\right|} {w}_{k}^{t+1}.
\end{aligned}
\end{equation}

\begin{table*}[htb]
	\centering
	\caption{Top-1 classification accuracy(\%) on ImageNet-LT dataset with different FL methods, where the results are referred in \protect\cite{shang2022federated}. The best results are marked in bold.}
	\label{ImageNet}
	\begin{tabular}{llcccc}
		\hline \multirow{2}{*}{{\textbf{Type }}} & \multirow{2}{*}{{\textbf{Method}}} & \multicolumn{4}{c}{\textbf{ImageNet-LT}}  \\
		\cline { 3 - 6 } & & All & Many & Medium & Few \\
		\hline \multirow{5}{*}{\makecell[l]{{\text { Heterogeneity-oriented }} \\ \text { FL methods } }} & 
		\text { FedAvg } & 23.85 & 34.92 & 19.18 & 7.10 \\
		& \text { FedAvgM } & 22.57 & 33.93 & 18.55 & 6.73  \\
		& \text { FedProx } & 22.99 & 34.25 & 17.06 & 6.37  \\
		& \text { FedDF } & 21.63 & 31.78 & 15.52 & 4.48  \\
		& \text { CCVR } & 25.49 & 36.72 & 20.24 & 9.26  \\
		\hline \multirow{3}{*}{\makecell[l]{{\text { Imbalance-oriented }} \\ \text { FL methods } }} & \text { Fed-Focal Loss } & 21.60 & 31.74 & 15.77 & 5.52    \\
		& \text { Ratio Loss } & 24.31 & 36.33 & 18.14 & 7.41  \\
		& \text { FedAvg+$\tau$-norm} & 21.58 & 31.66 & 15.76 & 4.33 \\
		\hline \text { SOTA } & {\text{ CReFF}} & 26.31 & \textbf{37.44} & 21.87 & 10.29  \\
		\hline
		\hline \multirow{2}{*}{\text { Proposed method } }& \multirow{2}{*}{\text{ CLIP2FL}}  & \textbf{27.72} & 35.06 & \textbf{27.55} & \textbf{24.80}  \\
		&\text {$\quad\quad$ } 
        & ($ \textcolor{red}{\uparrow 1.41}$) 
        & ($ \textcolor{green}{ \downarrow 2.38 }$)
        & ($ \textcolor{red}{\uparrow 5.68}$)  
        & ($ \textcolor{red}{\uparrow 14.51}$)\\
		\hline
	\end{tabular}
\end{table*}

 The retraining model is the copy of the global model, which is retrained by fixing the feature extractor, while the server classifier $g_{\varphi}$ w.r.t ${\hat{\varphi}^t}$ is updated with federated features and get the new parameter ${\hat{\varphi}^{t+1}}$ for round $t+1$.
\begin{equation}
\begin{aligned}
{\hat{\varphi}}^{t+1} \leftarrow {\hat{\varphi}}^{t}-\eta \nabla_{{\hat{\varphi}}} L_{ce}\left({v}_{i}^{t}, y_{i}\right).
\end{aligned}
\end{equation}

Finally, we obtain the re-trained model ${\hat{w}}^{t+1} = \{{\theta}^{t+1}, {\hat{\varphi}}^{t+1}\}$ and the updated global model ${{w}}^{t+1} = \{{\theta}^{t+1}, {{\varphi}}^{t+1}\}$. These models are then sent to clients for the next round. The comprehensive training process is presented in Algorithm \ref{algorithm}.

\begin{algorithm}[tb]
	\caption{Training Process for Round $t$}
	\label{algorithm}
	\textbf{Input}: global model parameter ${w}^t=\{{\theta}^t, {\varphi}^t\}$ and current re-trained classifier  ${\hat{\varphi}}^t$ , randomly initialized federated features $\left\{{{V}}_{c,i}^t \mid c \in {C}^{k}\right\}|_{i=1}^m$\\
	\textbf{Output}: a re-trained model parameterized as ${\hat{w}}^{t+1}=\{{\theta}^{t+1}, {\hat{\varphi}}^{t+1}\}$ for next round

	\begin{algorithmic}[1] 
		\STATE {\textcolor{gray} {\# Server executes:}}
		\STATE Send ${w}^t$ to clients;
		\STATE Randomly select a set of online clients $\Omega^t$;
		
		\STATE  {\textcolor{gray} {\# Client executes:}}\;
		
		\FOR{$k \in \Omega^t$}
		\STATE Update local model ${w}_{k}^{t+1}$ by Eq.\ref{local_training_loss} and Eq.\ref{update_local_model};
		\STATE compute real feature gradients $\left\{g_c^k \mid c \in {C}^{k}\right\}$ using Eq.\ref{compute_real_gradient};
		\ENDFOR
		\STATE Send $\left\{{w}_{k}^{t+1}, g^k \mid k \in \Omega^t\right\}$ to the server;
		
		\STATE  {\textcolor{gray} {\# Server executes:}}\;
		\STATE Update the global model ${w}^{t+1}$ by Eq.\ref{update_global_model};
		\STATE Aggregate real feature gradients to ${g}_{c}^{{agg}}$ using Eq.\ref{aggregate_real_gradient};
		\STATE Compute federated feature gradients ${g}_{c}^{{v}}$ by Eq.\ref{compute_virtual_gradient}
		\STATE Optimize federated features to ${{V}}^{t+1}$ by
		 gradient matching loss and prototype contrastive learning loss (Eq.\ref{loss_grad}, Eq.\ref{loss_ctra} and Eq.\ref{optimition_virtual_feature_total_loss})\;
		\STATE Re-train the classifier  ${\hat{\varphi}^t}$ to  ${\hat{\varphi}^{t+1}}$
		\STATE Send ${{w}}^{t+1}=\{{\theta}^{t+1}, {{\varphi}}^{t+1}\}$ and ${\hat{w}}^{t+1}=\{{\theta}^{t+1}, {\hat{\varphi}}^{t+1}\}$ to clients.
	\end{algorithmic}
\end{algorithm}
\section{Experiment}
\subsection{Experiment Setup}
\textbf{Datasets.} We implement CLIP2FL on three frequently used datasets with the long-tailed data:
CIFAR-10/100-LT~\cite{krizhevsky2009learning} and ImageNet-LT~\cite{russakovsky2015imagenet}. As for the first two datasets, we follow the previous work~\cite{cui2019class} to sample these two datasets with different imbalance factors (IF = 100, 50, 10) for long-tailed distribution, and we follow CReFF~\cite{shang2022federated} to use Dirichlet distribution with the key parameter $\alpha$ to generate the heterogeneous data partition among clients, where the value of $\alpha$ is set to 0.5 on CIFAR-10/100-LT. ImageNet-LT~\cite{russakovsky2015imagenet} has 115.8K images from 1000 classes and the number of images per class ranging from 1280 to 5, where the value of $\alpha$ is set to 0.1.
~\\\textbf{Implementation.} ResNet-8 is used as the feature extractor on  CIFAR-10/100-LT and ResNet-50 is used for ImageNet-LT. Note that in order to work with CLIP, a layer of MLP is added to both ResNet-8 and ResNet-50 to make their features have the same dimension as the output of CLIP. CLIP with the ViT-B/32 model is used in our experiments. 

The number of clients is set to 20, and  40\% of them are randomly selected as online clients to participate in training. The batch size of client-side training is set to 32 for all datasets and we set the number of federated features to 100 for each class. Experiments were conducted using PyTorch on four NVIDIA GeForce RTX 3090 GPUs. We employed the standard cross-entropy loss by default and executed 200 communication rounds.

\subsection{Experiment Results}

We compare CLIP2FL with eleven FL methods: FedAvg~\cite{mcmahan2017communication}, FedAvgM~\cite{hsu2019measuring}, FedProx~\cite{li2020federated}, FedDF~\cite{lin2020ensemble}, FedBE~\cite{chen2020fedbe}, CCVR~\cite{luo2021no} and FedNova~\cite{wang2020tackling}, Fed-Focal Loss~\cite{sarkar2020fed}, Ratio Loss~\cite{wang2021addressing} and FedAvg with $\tau$-norm~\cite{kang2019decoupling}, CReFF~\cite{shang2022federated}. The first seven methods are heterogeneity-oriented methods, and Fed-Focal Loss~\cite{sarkar2020fed}, Ratio Loss~\cite{wang2021addressing} and FedAvg with $\tau$-norm~\cite{kang2019decoupling} are imbalance-oriented methods.\\ 
\textbf{Results on CIFAR-10/100-LT.}
In Table \ref{CIFAR10-100}, we compare the classification accuracy of our CLIP2FL with various FL methods on CIFAR-10-LT and CIFAR-100-LT. It is observed that CLIP2FL achieves the best classification accuracy on both  CIFAR-10-LT and CIFAR-100-LT. When IF = 50, CLIP2FL gains 2.27\% on CIFAR-10-LT  and 3.65\% on CIFAR-100-LT in classification accuracy compared with CReFF, respectively. Even when IF = 100 which causes the data to be severely imbalanced, CLIP2FL achieves the gain by nearly 3\%. It fully demonstrates the superiority and effectiveness of CLIP2FL.\\
\textbf{Results on ImageNet-LT.} 
In Table \ref{ImageNet}, we show the classification accuracy comparison of CLIP2FL with different FL methods on ImageNet-LT. As shown in Table \ref{ImageNet}, the four divisions are defined as follows: ``Many" ($>100$ samples), ``Medium" ($20-100$ samples), ``Few" ($<20$ samples), and ``All" (overall accuracy), respectively. It is observed that CLIP2FL is superior to the compared methods in the accuracy of the above four divisions, especially CLIP2FL achieves a significant gain of 5.68\% and 14.51\% on the ``Medium" and ``Few" divisions compared with CReFF.
It indicates that CLIP2FL not only enhances the overall performance but also obviously improves the classification performance of the tail categories. The comparison on ImageNet-LT further highlights the advantages of CLIP2FL, especially when dealing with long-tailed datasets.

\subsection{Model Validation}
\textbf{Ablation Study.} We conduct ablation studies with two losses on CIFAR-10/100-LT: 1) The knowledge distillation loss from CLIP to the client-side model (defined as $KL$); 2) The prototype contrastive learning loss on the server side for generating federated features (defined as $L_{pcl}$). We select the method~\cite{shang2022federated} as our baseline. As shown in Table \ref{ablation}, each component is revealed and plays an important role in our CLIP2FL. CLIP2FL only with $KL$ is superior to the baseline method with the gain of 1.65\% and 1.21\% on CIFAR-10/100-LT, which shows CLIP can help the client-side model to improve the ability of feature representation. At the same time, the baseline method with the two losses achieves the gain of 2.82\% and 2.89\% on CIFAR-10/100-LT. It shows that CLIP benefits to supervise the generation of federated features. 
In addition, this result shows that $KL$ and $L_{pcl}$ can complement each other, which is consistent with our motivation.

\begin{figure}[tbp]
	\centering
	\includegraphics[width=1.0\linewidth]{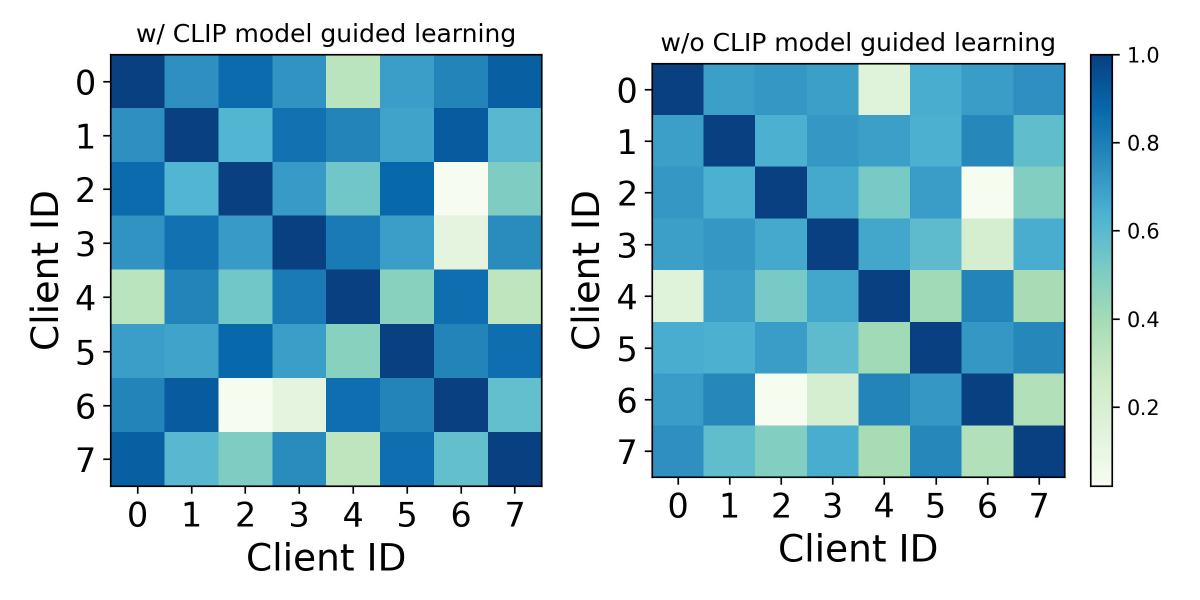}
	\caption{Visualization of CKA similarity at the classifier layer under the ``client model-client model" setting with or without using the CLIP model to guide the client model.}
	\label{clip_distillation}
\end{figure}

~\\\textbf{Visualization.} In order to further verify that CLIP benefits to guide the client learning and to alleviate classifier bias caused by long-tailed data, we follow the CCVR~\cite{luo2021no} and adopt Centered Kernel Alignment (CKA)~\cite{kornblith2019similarity} to measure the similarity of the output features between two local client models given the same input. CKA outputs a similarity score between 0 (not similar at all) and 1 (identical). The visualization results are shown in Figure \ref{clip_distillation}, the darker color indicates a higher degree of similarity. Compared with the w/o CLIP model (right), the CLIP-guided client model (left) significantly improves the CKA similarity between different client models, it demonstrates that using the CLIP model to guide client learning can effectively reduce the bias of the classifier, which means improving feature representation learning more fully.
\setlength{\tabcolsep}{0.9mm}{
	\begin{table}[tbp]
		\renewcommand\arraystretch{1.2}
		\centering
		\caption{Ablation study to investigate the effectiveness of CLIP2FL on CIFAR-10-LT and CIFAR-100-LT with IF = 100.}
		\label{ablation}
		\begin{tabular}{c|cc|cc}
			\toprule[1pt]
			Method & $KL$   & $L_{pcl}$ & CIFAR-10-LT & CIFAR-100-LT \\ \hline
			  Baseline &  &  & 70.55  & 34.67     \\ \hline
			  w/o $L_{pcl}$ & \checkmark &   & 72.20  & 35.88 \\ \hline
			CLIP2FL  & \checkmark  & \checkmark & \textbf{73.37} & \textbf{37.56}  \\
			\bottomrule[1pt]
		\end{tabular}
	\end{table}
}
~\\\textbf{Hyperparameter Analysis.} Three important hyperparameters in our CLIP2FL are $\beta$, $\eta$ and $m$. $\beta$ controls the distillation weight of the CLIP model to guide the client model's training, $\eta$ controls the balance of prototype contrastive learning loss and gradient matching loss when optimizing federated features, and $m$ is the number of federated features initialized for each class. We analyzed hyperparameters under the condition that only one hyperparameter is selected as the variable, while all other hyperparameters are kept constant. We found that CLIP2FL achieved the best performance when $\beta$ = 3.0, $\eta$ ${\in}$  $\{0.001, 0.0001, 1e-5\}$ and $m$ = 100. Surprisingly, we observed that the performance of CLIP2FL exhibits limited sensitivity to the hyperparameters, indicating a degree of robustness in the CLIP2FL. More details are in the Appendix.

\begin{figure}[t]
	\centering
	\includegraphics[width=1.0\linewidth]{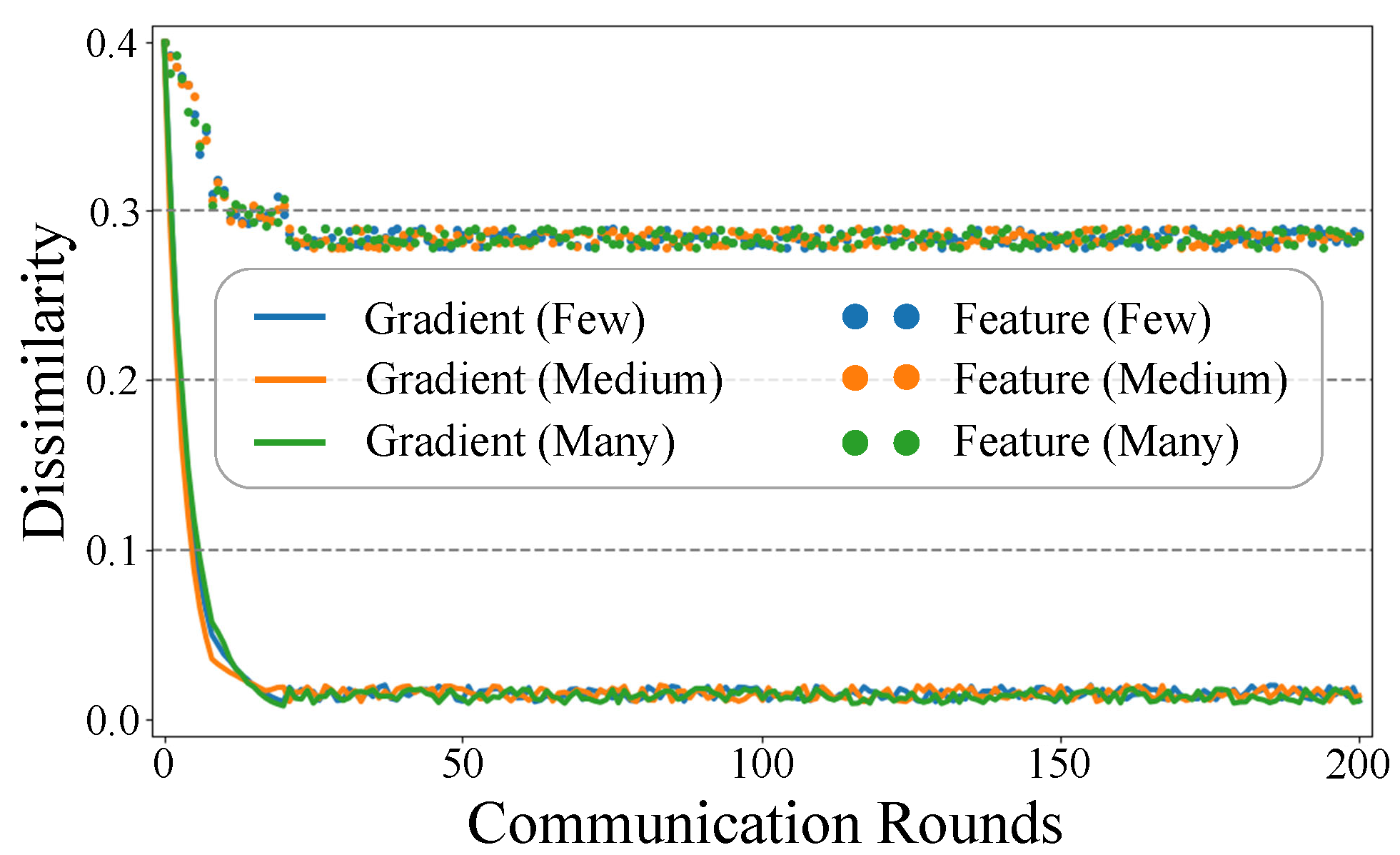} 
	\caption{Analysis of two dissimilarities on CIFAR-10-LT with IF = 100. The dotted lines denote the dissimilarities between federated features and real features, and the solid lines denote the dissimilarities between federated feature gradients and real feature gradients.}
	\label{gradient_feature}
\end{figure}

\subsection{Dissimilarity Analysis of Gradient \& Feature}
We discuss how the dissimilarity between the federated features gradients and the real sample features gradients change with the increase of iterations (Eq. \ref{optimition_virtual_feature_total_loss}). Here, the classes of CIFAR-10-LT are divided into three groups: Many ($>1500$ samples), Medium ($200-1500$ samples) and Few ($<200$ samples), and the average dissimilarities for each group are calculated separately. As the solid lines shown in Figure \ref{gradient_feature}, after several rounds of communication, the dissimilarities between the two gradients tend to an approximate value of 0. It shows that the loss can successfully optimize the gradients generated by the federated features to be consistent with the gradients of the real sample features. It indicates that the retraining classifier with federated features on the server side is effective. Moreover, we also calculate cosine dissimilarity between the federated features and the real sample features, which is shown in the dotted lines in Figure \ref{gradient_feature}. With the increase of training rounds, the dissimilarities between federated features and real sample features are also decreasing but are larger than 0. It demonstrates that although the corresponding gradients are very close, the generated federated features are not completely similar to the real sample features, which ensures the privacy of real sample features.

\section{Conclusion}
In this work, we provide the CLIP-guided FL method (CLIP2FL) for federated learning on heterogeneity and long-tailed data. On the client side, a ready-made CLIP model is used for knowledge distillation to the client models. On the server side, CLIP is used to guide the generation of federated features via prototype contrastive learning. The federated features are generated to mitigate the strong bias problem caused by long-tailed data. CLIP2FL enforces the semantic supervision on FL and bridges the client-side and server-side training with CLIP. The experimental results on three benchmarks achieve a significant superiority against the state-of-the-art methods. 

\section{Acknowledgments}
This work is supported by the National Natural Science Foundation of China under Grants (62176224, 62222602, 62176092, 62376233), Natural Science Foundation of Chongqing (CSTB2023NSCOJOX0007), CCF-Lenovo Blue Ocean Research Fund.

\bibliography{aaai24}

\end{document}